# GNN-ASE: Graph-Based Anomaly Detection and Severity Estimation in Three-Phase Induction Machines

Moutaz Bellah Bentrad[1*] , Adel Ghoggal[1] , Tahar Bahi[2,3] , Abderaouf Bahi[4]

[1]Electrical engineering laboratory of Biskra (LGEB) Mohamed Khider University, BP 145 RP 07000 Biskra, Algeria.
[2]Electrical Engineering Department, Laboratory of Automation and Signals of Annaba (LASA)
[3]Badji Mokhtar-Annaba University,P.O Box.12, Annaba, 23000 Algeria
[4]Computer Science and Applied Mathematics Laboratory, Chadli Bendjedid El Tarf University, El Tarf 36000, Algeria

moutazbellah.bentrad@univ-biskra.dz*, a.ghoggal@univ-biskra.dz, tahar.bahi@univ-annaba.dz, a.bahi@univ-eltarf.dz

**Abstract**
The diagnosis of induction machines has traditionally relied on model-based methods that require the development of complex dynamic models, making them difficult to implement and computationally expensive. To overcome these limitations, this paper proposes a model-free approach using Graph Neural Networks (GNNs) for fault diagnosis in induction machines. The focus is on detecting multiple fault types—including eccentricity, bearing defects, and broken rotor bars—under varying severity levels and load conditions. Unlike traditional approaches, raw current and vibration signals are used as direct inputs, eliminating the need for signal preprocessing or manual feature extraction. The proposed GNN-ASE model automatically learns and extracts relevant features from raw inputs, leveraging the graph structure to capture complex relationships between signal types and fault patterns. It is evaluated for both individual fault detection and multi-class classification of combined fault conditions. Experimental results demonstrate the effectiveness of the proposed model, achieving 92.5% accuracy for eccentricity defects, 91.2% for bearing faults, and 93.1% for broken rotor bar detection. These findings highlight the model's robustness and generalization capability across different operational scenarios. The proposed GNN-based framework offers a lightweight yet powerful solution that simplifies implementation while maintaining high diagnostic performance. It stands as a promising alternative to conventional model-based diagnostic techniques for real-world induction machine monitoring and predictive maintenance.

## 1.Introduction

Electrical rotating machinery play a pivotal role in the industrial chain and have considerably contributed economically due to their efficiency and the demand on them is reported to reach 30% in the industry by 2040 [1,2], Induction machines (IMs) and though their robustness and reliability may suffer from critical sudden stoppages that may influence their performance [3-5]. Therefore, preventive maintenance has become a necessity now more than ever to continually assess the health state of the IM to make sure that the mechanism functions properly.

The electrical and mechanical systems involved in the IM produce stresses by the defects that happen during operation and may result in damages of different parts of the machine.among these faults rotor, bearing parts and stator related defects can be found [6-8], these defects represent approximately 32% for stator related faults, 10% are for the rotoric defects and 40% represent bearing related stresses [9,10], the occurrence of faults is inherent to its operation, even under ideal operating conditions [11,12].

In general, to conduct a typical intelligent diagnosis of the machine, following phases must be taken into account: data acquisition, constructing a robust mode, feature extraction and finally decision making [13]. For data collection for both healthy and faulty data, they layout of sensors is vital for capturing important details about the health condition and reduces interference of transfer paths [14].The main focus of of research is on developing multi source signal samples dedicated for training, for instance a conditional variational generative adversarial network [15] which would be based on data fusion as an addition for fault samples using current, vibration and acoustic signals simultaneously [16,17]

Induction machine faults have been diagnosed using various frequency analysis methods, including fast Fourier transform (FFT), power spectrum estimation, and envelope spectrum analysis [18]. The disadvantage of this technique is based on the fact that it requires that the signal being processed is stationary and lineare and that the rotational speed has transient mode, which represents a limitation by this technique. Therefore, to mitigate this obstacle a Time-Frequency analysis is performed such as Short Fast Fourier Transform (STFT) [19], the premise of this technique is based on the FFT method with applying a sliding window to enhance the frequency resolution. However, the limitation of this method is the fixed size of the window which is not practical with the transient nature of the signal. Another method is being applied for this regard, Wigner-Ville distribution (WVD) [20] which delivers better frequency resolution under the same conditions, another drawback of this approach is the generation of undesirable large frequencies. Empirical mode decomposition (EMD) [21-23] on the other hand, is a self-adaptive approach, which produce a series of intrinsic mode functions (IMFs) and with the help of the Hilbert Transform (HT) [24], it can estimate instantaneous frequencies, but its limitation resides on the generation of undesirable negative frequencies. Wavelet Transform (WT) delivers enhanced frequency resolution results thanks to its local frequency-time domain properties [25].

Intelligent online diagnosis can be employed since AI-driven approaches presentend in Figure 1, proved reliable against traditional techniques by reducing interaction with the machine, these approaches rely only on provided data which means no extensive knowledge of the machine is required, they use association and human like reasoning and finally decision making, they use signal processing and classification by employing Machine learning (ML) like the artificial neural network (ANN), fuzzy logic (FL), fuzzy neural network (FNN), genetic algorithm (GA), Hidden Morkov model, Bayesian classifier, Support Vector Machine (SVM), Deep Learning (DL) [26-29].

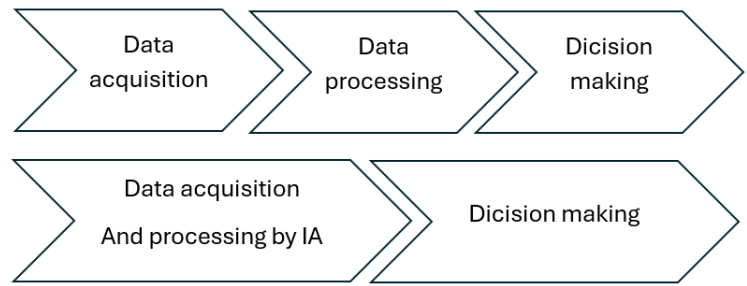

**Fig. 1 Diagnosis approaches using IA**

The primary objective of this study is to explore the effectiveness of the GNN-ASE algorithm in detecting and classifying defects such as eccentricity, broken bars, and bearing faults in induction machines. This research aims to use both current and vibration raw signals to detect these defects without the need for traditional signal preprocessing, thanks to the GNN-ASE's advanced feature extraction capabilities, which can be resumed by:

- Automated Feature Extraction: by leveraging GNN-ASE's ability to automatically detect complex relationships and patterns within raw signals, this study eliminates the need for conventional preprocessing techniques (e.g., filtering, Fourier transforms). This offers a more direct, data-driven approach to diagnosing defects, allowing for the identification of subtle fluctuations in signal patterns that may indicate the early stages of failure. The algorithm's capacity to extract meaningful features directly from raw data simplifies the diagnostic process and increases adaptability across various defect types.

- Comprehensive Health Monitoring: the second goal is to investigate how GNN-ASE can assess the health of the induction machine under various operating conditions and severity of defects. By analyzing the raw signal data at different severity levels, the study aims to establish a robust framework for real-time health monitoring, providing deeper insights into machine performance degradation. This will aid in precision maintenance planning, allowing for proactive interventions that can prevent unexpected breakdowns, thereby improving machine reliability and reducing downtime.

- Defect analysis and characterisation : the research will offer insights into the nature of each fault and how it evolves over time. This will help establish a clearer understanding of the failure modes in induction machines, contributing to the body of knowledge on machine fault diagnosis.

## 2. Related works

The diagnosis of faults in induction machines has been a long-standing topic of interest in the field of condition monitoring and predictive maintenance. Numerous techniques have been explored over the years, ranging from model-based approaches to data-driven methods that rely on machine learning and signal processing. This section presents a review of key contributions in the literature, highlighting recent efforts that focus on fault detection using hybrid signals, advanced neural networks, and statistical feature extraction. The following works illustrate the diversity of strategies used to detect faults such as bearing defects, broken rotor bars, and other abnormalities in induction machines.

Saucedo et al. in [30] suggested to monitor the health condition of the induction machine by inspecting bearing defect case with various severities and proposed a method by fusing the current-vibration signal gathered from healthy and faulty machines and by employing machine learning uses a hybrid dataset of statistical features which would be prepared by applying a time-domain and a frequency-domain analysis, the classifier model would later on validate the efficiency of the suggested approach.

Marmoush et al. in [31] proposed using radial basis function neural networks (RBFN) and probabilistic neural networks (PNN) to apply an online diagnosis of the induction machine for rotor fault classification, and by reducing the dimensionality of the provided dataset through principal component analysis that improved significantly the learning speed and the cost of data processing.

Min-Chan et al. in [32] suggested applying SVM, DL, Convolutional Neural Network, Gradient Boosting and Xgboost for the diagnosis of rotor defect and bearing faults by simulating these different faults under various conditions using a simulator for data collection, then experimental results validate the suggested approach and its effectiveness.

Defdaf et al. in [33] proposed using preprocessing the signal collected from three-phase induction machine through discrete wavelet transform (DWT) to extract features related to broken bars defect and by applying two statistical studies which are Root Mean Square (RMS) and Kurtosis shock factors values, these features will be used as inputs to train an ANN model, the latter delivered an accuracy of prediction that reaches 98.66%.

In Table 1 a recapitulation for previous works is shown, which includes the advantages, limitations, and approach used. Despite the great advancement in anomaly detections, some gaps still exist in recent research, GNN-ASE fills the discovered gaps by showing encouraging results in terms of Accuracy, Recall and F1-Score

Table 1. State of the art summary

| Ref | Year | Approach | Advantages | Limitations |
|---|---|---|---|---|
| [30] | 2021 | multilayer NNs based classifier | High-performance signal characterisation, accurate classification ability | Need for signal processing by performing Time and Frequency domain analysis. |
| [31] | 2017 | RBFN, PNN | Reduced data process time and speed learning | Applying additional Principal Component Analysis for dimensionality reduction |
| [32] | 2023 | SVM, multilayer NNs, CNN, gradient boosting, and XGBoost | Fast and accurate diagnosis of rotor and bearing fault | Lack of severity degree classification |
| [33] | 2021 | ANN | High classification rate for broken rotor bars | applying time and frequency domain analysis and statistical selection for feature extraction |

## 3. Defects of the induction machine

### 3.1 Air gap irregularity

Eccentricity is a prevalent mechanical issue in induction motors, as noted in [34]. It can manifest in three forms: static, dynamic, or a combination of both, as depicted in Figure 2. Static eccentricity, a key subtype, typically arises from ovality in the stator core or misalignment between the rotor and stator during initial installation. In contrast, dynamic eccentricity emerges from worn bearings, mechanical resonance, or a bent rotor shaft, leading to rotational misalignment. These factors together distort the air gap, causing fluctuations in both air-gap voltage and line current, ultimately destabilizing motor performance.

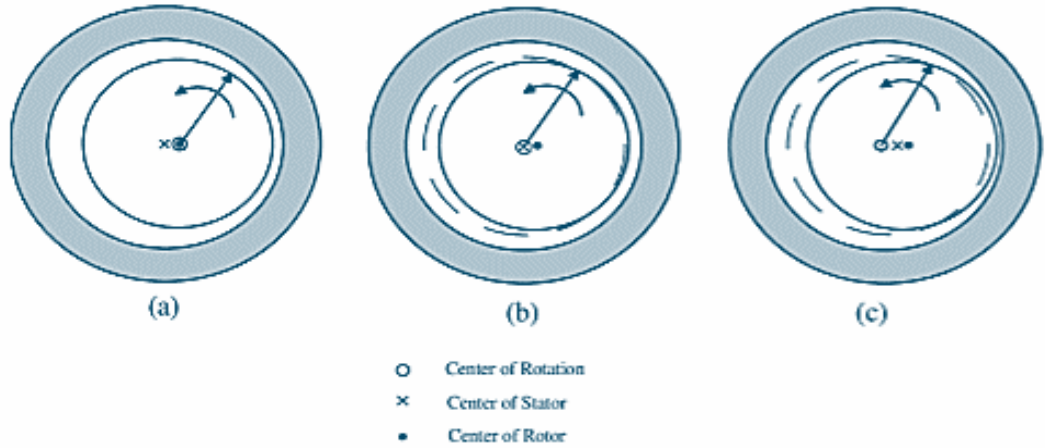

Fig. 2 Air gap eccentricity types : a- Static eccentricity, b- Dynamic, c- Mixed

Figure 3 shows clearly the impact of air gap irregularity on stator current across all three phases by causing harmonics which would distort the sinusoidal shape of the current and causes fluctuations leading to losses in the power supply and decrease in the motor's efficiency, the following are stator current signals collected from simulating air gap irregularity defect for healthy and defected machines at early and extreme cases.

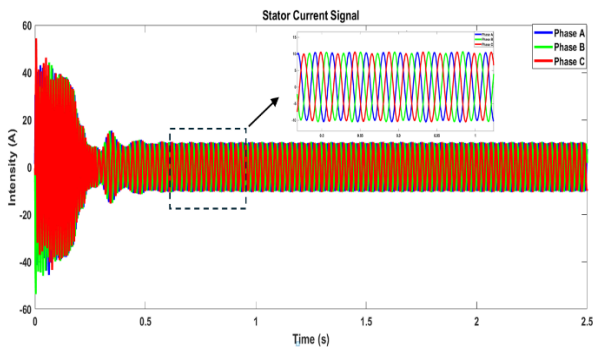 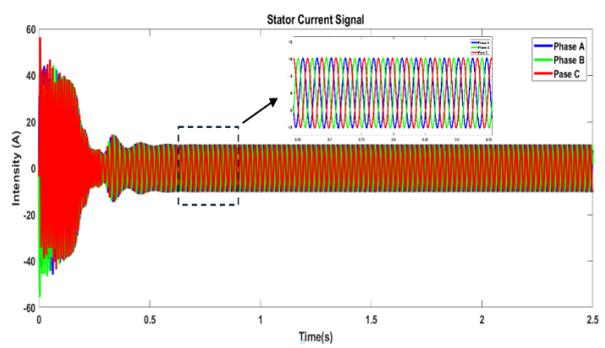

(a) Healthy stator current

(b) Air gap irregularity early stage

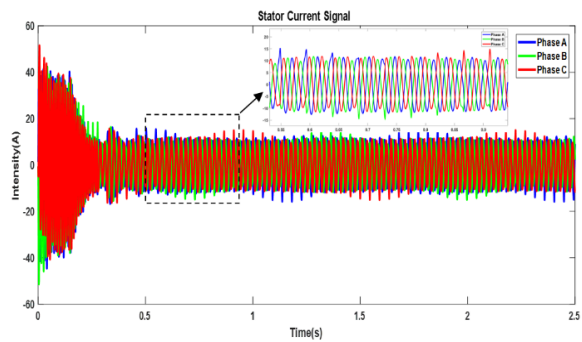

(c) Air gap irregularity extreme stage

Fig. 3 Air gap eccentricity influence on stator current

## 3.2 Broken rotor bars

The presence of broken rotor bars in a squirrel cage induction motor (IM) creates asymmetry in the rotor current distribution since no current flows through the damaged bars [35]. This disruption alters the air gap magnetic field distribution, which in turn induces voltages in the stator windings. Consequently, currents flow through both the windings and the power supply. The forward and backward rotating magnetic fields in the air gap generate line currents at frequencies of $(1 - 2s)f_s$ and $(1 + 2s)f_s$, respectively. where $s$ is the slip, and $f_s$ is the supply frequency. Broken rotor [36] bars can be identified by Formula (1).

$$f_{brb} = f_s \left[ k\left(\frac{1-s}{p}\right) \pm s \right] \quad (1)$$

When a rotor bar first breaks, the resistance between the broken bar and the adjacent healthy bars remains low, allowing current to continue flowing into the broken bar [37]. The current enters the broken rotor bar (BRB) from the healthy side, travels along the bar's length, and exits through the laminated core into the neighboring healthy bar.

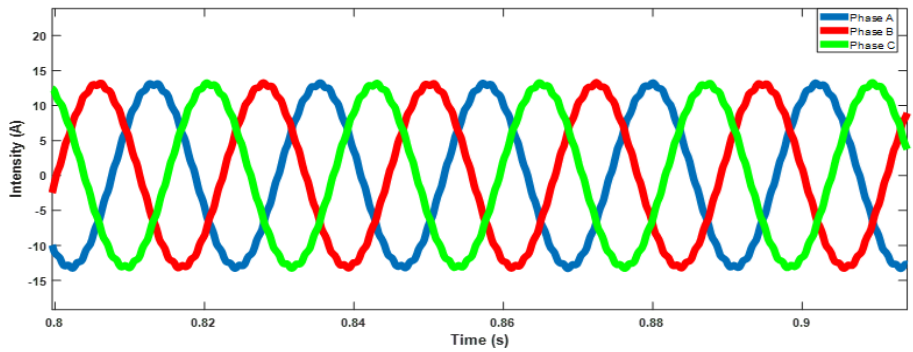

**(a) Healthy signal**

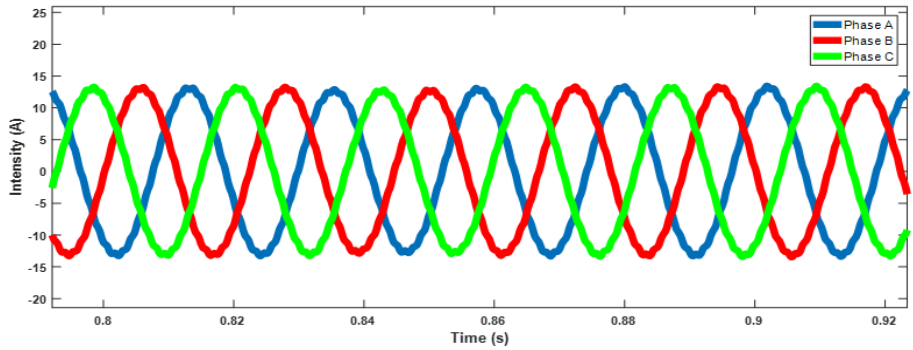

**(b) one broken bar**

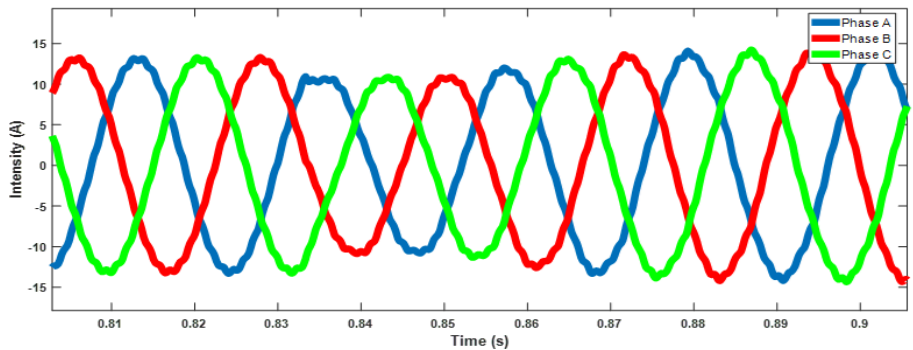

**(c) three broken bars**

**Fig. 4 Stator current signal of broken rotor bars fault**

## 2.3 Bearing defect

Bearing faults are the most common faults in induction machines (IMs), as discussed earlier. Rolling bearings consist of an inner race, an outer race, and cylindrical rollers or balls in between. Damage like flaking and pitting can occur due to material fatigue or wear in any of these parts [38]. In our analysis presented in Figure 5, we focus on two types of bearing faults: inner race and outer race faults. Bearing damage causes radial motion between the stator and rotor, leading to oscillations that induce characteristic fault frequencies in the current signals. This is due to the radial displacement of the rotor, which alters the machine's inductance and causes amplitude, frequency, and phase modulation in the motor current [39,40]. Expressed in Formula (2).

$$i(t) = \sum_{k=1}^{\infty} ik.cos(\omega_{Ck}.t + \varphi) \quad (2)$$

where $\omega_{Ck}$ and $\varphi$ represent angular velocity and phase angle respectively $\omega_{Ck} = \frac{2\pi f_{Bearing}}{P}$ and $f_{Bearing}$ is the bearing current harmonic frequency represented in Formula (3).

$$f_{Bearing} = |f_s \pm m f_v| \quad (3)$$

$f_s$ is the supply frequency, m=1,2,3… represent harmonic index, $f_v$ denotes fault's type frequency (ball, inner race, outer race)

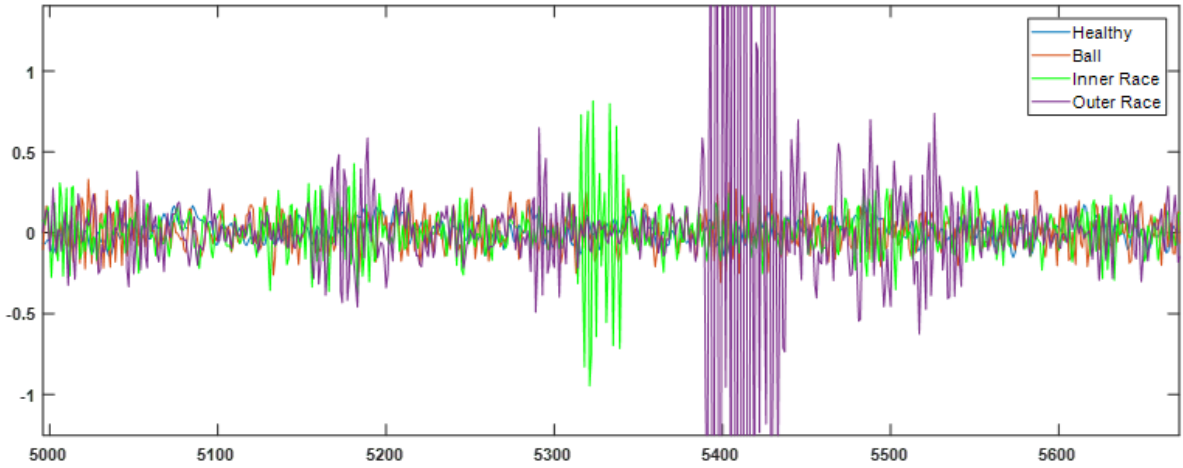

**Fig. 5 Bibrational signal for bearing defect**

## 3. Methodology

This section provides a detailed description of GNN-ASE and presents an overview of the proposed model's workflow.

### 3.1 Data Preprocessing

Prior to constructing the graph and implementing the GNN-ASE model, the raw electrical signal data necessitates preprocessing. This stage entails cleansing, standardizing, and augmenting the data to guarantee the model's resilience and to enhance its ability to identify abnormalities.

Electrical signals frequently exhibit noise attributable to sensor deficiencies or ambient influences. To maintain the critical elements of the signal, we implement a filtering technique, such as a Butterworth filter, to eliminate extraneous noise and mitigate the likelihood of false positives in anomaly identification. The filtered signal, $S_{clean}(t)$, is derived by applying the filter $H(f)$ to the raw signal $S_{raw}(t)$ in Formula (4).

$$S_{clean}(t) = F^{-1}(H(f)F(S_{raw}(t))) \quad (4)$$

where $F$ and $F^{-1}$ denote the Fourier and inverse Fourier transforms, and $H(f)$ is the frequency response of the filter.

In anomaly detection, datasets are often imbalanced, as normal signals typically outnumber anomalous ones. To address this, data augmentation techniques are applied to increase the diversity of the training data and ensure better generalization. The following augmentation techniques are used:

Time-shifting: A slight shift in the signal along the time axis creates new variations, making the model more robust to time-aligned differences. If the original signal is $S(t)$, the augmented signal $S_{shifted}(t)$ can be generated by shifting it by a time interval $\Delta t$. This is represented in Formula (5).

$$S_{shifted}(t) = S(t + \Delta t) \quad (5)$$

Amplitude scaling: To simulate variations in the strength of the signal, amplitude scaling modifies the signal's amplitude by a factor αα, where αα is sampled from a uniform distribution within a specified range. The new signal $S_{scaled}(t)$ is given in Formula (6).

$$S_{scaled}(t) = \alpha S(t) \quad (6)$$

Noise addition: To diversify the dataset, low-level Gaussian noise is added to the original signal, resulting in the augmented signal $S_{noisy}(t)$ as represented in Formula (7).

$$S_{noisy}(t) = S(t) + \varepsilon(t) \quad (7)$$

where $\varepsilon(t) \sim N(0, \sigma^2)$ is Gaussian noise with mean 0 and variance $\sigma^2$.

Data augmentation ensures a more balanced and varied training set, which leads to better generalization and improves the model's ability to detect rare anomalies in electrical signals.

### 3.2 Feature Extraction

Feature extraction is a crucial step in transforming raw electrical signal data into meaningful inputs for the GNN-ASE model. The goal is to capture relevant time-domain and frequency-domain characteristics from the signals, enabling the GNN to effectively analyze and detect anomalies. In GNN-ASE, this feature extraction is managed by a dedicated hidden layer within the architecture, designed to automatically capture these features for both nodes and edges of the graph.

The time-domain features are directly derived from the raw signals and are crucial for capturing variations in the signal's amplitude and power. The following features are extracted:

Peak Amplitude: This is the maximum value of the signal over a given time window, denoted as $A_{peak}$, and is calculated as Formula (8).

$$A_{peak} = max(S(t)) \quad (8)$$

where $S(t)$ is the signal over time.

Root Mean Square (RMS): RMS measures the power of the signal and is calculated as the square root of the average of the squared signal values as represented in Formula (9).

$$RMS = \sqrt{\frac{1}{T}\sum_{t=1}^{T} S(t)^2} \quad (9)$$

where $T$ is the total number of samples in the time window.

Signal Variance: This measures the variability in the signal's amplitude over time. The variance, $\sigma^2$, is calculated as Formula (10).

$$\sigma^2 = \frac{1}{T}\sum_{t=1}^{T}(S(t) - \mu)^2 \quad (10)$$

where $\mu$ is the mean value of the signal over the time window.

To capture frequency-based characteristics of the signal, we apply a Fast Fourier Transform (FFT), converting the time-domain signal into its frequency components. This enables the extraction of key frequency-domain features:

Dominant Frequency: The frequency at which the signal has the highest amplitude in the frequency spectrum. If $S(f)$ represents the signal in the frequency domain, the dominant frequency $f_{dom}$ is represented as Formula (11).

$$f_{dom} = argmax(S(f)) \quad (11)$$

Spectral Entropy: This measures the disorder or randomness in the frequency components, indicating the complexity of the signal. Spectral entropy $H_{spec}$ is calculated as Formula (12).

$$H_{spec} = \sum_{i=1}^{N} P(f_i)logP(f_i) \quad (12)$$

where $P(f_i)$ is the normalized power of the $i-th$ frequency component, and $N$ is the total number of frequency components.

These extracted features are used to represent the nodes (electrical signals) and edges (relationships between signals) in the graph. By incorporating both time-domain and frequency-domain features, the model gains a richer understanding of the signal's behavior, which enhances its ability to detect anomalies and assess their severity.

### 3.3 Graph Creation

The GNN-ASE model operates on graphs that represent the structure of the electrical signals. Each signal is transformed into a graph where nodes represent different segments or windows of the signal, and edges capture the relationships between these segments.

Each node in the graph represents a time window of the signal, characterized by the features extracted in Section 3.2. Formally, let $v_i$ be the $i-th$ node in the graph, with a feature vector $x_i$ containing the extracted features as represented in Formula (13).

$$v_i = \{x_i | x_i \in \Re^d\} \quad (13)$$

where $d$ is the dimensionality of the feature vector. The edges between nodes represent the temporal and frequency relationships between consecutive or correlated signal segments. Each edge $e_{ij}$ between nodes $v_i$ and $v_j$ is weighted according to the similarity between the feature vectors of the connected nodes, with weights initialized based on cosine similarity as represented in Formula (14).

$$w_{ij} = \frac{x_i x_j}{\|x_i\|\|x_j\|} \quad (14)$$

After graph initialization, the edges are dynamically reweighted during the training process using the dynamic edge reweighting mechanism.

*3.4 GNN-ASE Implementation & Training*

The GNN-ASE architecture is designed to dynamically update edge weights and incorporate an exploration-exploitation balance while learning to detect anomalies and estimate their severity. The main components of the GNN-ASE model are outlined below.

The GNN-ASE begins with two Graph Convolutional Network (GCN) layers that extract and propagate the node features across the graph, allowing each node to aggregate information from its neighbors. Formally, for each node $v_i$, the hidden representation at the $l^{th}$ layer is computed as Formula (15).

$$h_i^{(l+1)} = \sigma\left(\sum_{j \in N(i)} \frac{w_{ij}}{\sqrt{d_i d_j}} h_j^{(l)} W^{(l)}\right) \quad (15)$$

where $h_i^{(l)}$ is the hidden representation of node $i$ at layer $l$, $N(i)$ is the set of neighboring nodes of $v_i$, $w_{ij}$ is the edge weight between nodes $i$ and $j$, $d_i$ is the degree of node $i$, $W^{(l)}$ is the weight matrix for layer $l$, $\sigma$ is an activation function.

One of the key innovations in GNN-ASE is the dynamic edge reweighting mechanism, which adjusts the edge weights during the recommendation process to promote diversity and better exploration of the graph structure.

At each step, the edge weights $w_{ij}$ are updated based on user interactions or signal changes using the following update rule as represented in Formula (16).

$$w_{ij}^{new} = (1-\beta) w_{ij}^{old} + \beta f(x_i, x_j) \quad (16)$$

where $w_{ij}^{old}$ is the previous edge weight, β is a hyperparameter controlling the reweighting degree ($0 \leq \beta \leq 1$), $f(x_i, x_j)$ is a function that measures the interaction or similarity between nodes.

The dynamic edge reweighting ensures that the model adapts to new data patterns while retaining learned relationships, helping the model explore alternative hypotheses in anomaly detection.

To estimate the severity of each detected anomaly, we introduce a severity estimation layer, which takes as input the final node embeddings from the GCN layers and outputs a severity score $s_i$ for each node $v_i$ as represented in Formula (17).

$$s_i = Softmax(W_{severity} h_i^{(L)}) \quad (17)$$

where $W_{severity}$ is a weight matrix and $h_i^{(L)}$ is the final hidden representation of node $i$. The severity score reflects how critical the anomaly is, with higher values indicating more severe issues in the electrical signal.

Finally, the output layer produces three outputs for each node:
- Anomaly detection: A binary output indicating whether the node represents an anomaly.
- Severity score: A continuous score indicating the severity of the detected anomaly.
- Anomaly type: The output layer also classifies the type of anomaly detected.

These outputs are combined to provide both the detection and evaluation of anomalies in the electrical signals.

**4. Experimental evaluation**

*4.1 Datasets*

Air gap eccentricity and broken rotor bars dataset consists of raw stator current signals from all three phases of a simulated three-phase squirrel cage induction machine. These signals were collected under various fault types and load conditions to diagnose. For bearing defects, vibrational signals were gathered, the detailed data collection is illustrated Table 2.

**Table 2. Dataset collection**

| Machine Condition | Type | Load condition | Severity |
|---|---|---|---|
| Healthy | — | 0/10/30/40 N.m | — |
| Air gap eccentricity | Static | 0/10/30/40 N.m | 10%-20%-30%-40% |
|  | Dynamic | 0/10/30/40 N.m |  |
|  | Mixed | 0/10/30/40 N.m |  |
| Broken rotor bars | — | 0/10/30/40 N.m | 1 BRB-2 BRB-3 BRB |
| Bearing | Ball | No load/weak/medium/heavy | 7 inch-14 inch-21 inch |
|  | Inner race |  |  |
|  | Outer race |  |  |

*4.2 Evaluation Metrics*

We chose to evaluate GNN-ASE, three important and well-known evaluation metrics in the field of machine learning, allowing for comparison with other cutting-edge approaches.

Accuracy: We employ accuracy as a general measure of performance. It is calculated by determining the proportion of correct predictions (both true positives and true negatives) relative to the total number of predictions made. Formula (18) illustrates the mathematical expression for this metric.

$$Accuracy = \frac{TP + TN}{TP + TN + FP + FN} \quad (18)$$

Recall: Recall is essential for evaluating our model's capability to identify all relevant instances. Formula (19) provides the mathematical formulation of this metric.

$$Recall + \frac{TP}{TP + FN} \quad (19)$$

F1-Score: The F1-Score is the harmonic mean of precision and recall. Formula (20) represents the mathematical formulation of this metric.

$$F1 - Score = 2 \times \frac{Precision \times Recall}{Precision + Recall} \quad (20)$$

4.3 Model Architecture

To justify the choice of GNN-ASE's architecture inspired from MycGNN [41], we state that these values were selected after obtaining the best results through an iterative trial-and-error process. We tested multiple configurations by varying only one parameter at a time while keeping the others fixed. Ultimately, the optimal configuration is demonstrated in Table 3.

**Table 3. GNN-ASE's architecture**

| Layer Type | Dimension | Activation Function | Hyperparameters |
|---|---|---|---|
| Input | Embedding nodes | 128 | - |
| Graph Convolutional Layer 1 | 64 | ReLU | Learning rate: 0.01 |
| Graph Convolutional Layer 2 | 64 | ReLU | Dropout: 0.5 |
| Edge Reweighting Layer | 64 | - | β (degree of reweighting) |
| Severity Evaluation Layer | 32 | ReLU | - |
| Output 1 (Anomaly Detection) | 1 | Sigmoid | - |
| Output 2 (Severity Score) | 1 | ReLU | - |
| Output 3 (Anomaly Type) | 3 | Softmax | classes: eccentricity fault, bar breakage, bearing fault |

*4.4 Ablation Study*

An ablation study helps evaluate the impact of different components of the GNN-ASE model by removing or altering one aspect at a time to observe its effect on the overall performance. This study aims to isolate key features such as edge reweighting, severity score evaluation, and feature extraction mechanisms to understand their contribution to the final performance of the model.

- **GNN-ASE@1:** In this experiment, the dynamic edge reweighting layer is removed. The model performance is compared to the full model, revealing how the edge adjustment mechanism contributes to both anomaly detection and diversity in recommendations. Accuracy and recall decrease slightly, particularly in detecting rare anomalies, as the edge reweighting layer enhances exploration by diversifying the recommendations.
- **GNN-ASE@2:** Here, the severity evaluation layer is omitted, and the model only focuses on binary anomaly detection. This helps assess how much additional value the severity estimation adds. While the model may retain its accuracy for binary anomaly detection, the ability to distinguish severe anomalies from less critical ones will be lost.
- **GNN-ASE@3:** The extraction of frequency-domain features, such as spectral entropy, is removed, leaving only time-domain features. The absence of frequency features reduces the model's ability to detect subtle anomalies that manifest more clearly in the frequency domain, leading to a reduction in both recall and F1-score.

This study confirms that each component plays a crucial role in enhancing the overall performance.

*4.5 Overview of GNN-ASE performance*

Table 4 above shows the performance of the GNN-ASE model and its variants across three datasets: Eccentricity Defects, Bearing Faults, and Bar Breakage Defects. Removing the edge reweighting in GNN-ASE@1 results in a noticeable drop in performance, with accuracy decreasing from 92.5% to 89.8% for Eccentricity Defects and similar declines across the other datasets. This shows the importance of dynamic edge reweighting in capturing complex relationships within the graph. However, in GNN-ASE@2, where severity estimation is excluded, the performance remains identical to the full model, as severity estimation only adds supplementary output without affecting the core anomaly detection capabilities. Finally, GNN-ASE@3, which omits frequency-domain features, shows reduced performance, particularly in the Bearing Faults dataset, where accuracy drops from 91.2% to 86.3%, demonstrating the value of including frequency features for more robust detection.

**Table 4. GNN-ASE's performance**

| Model | Dataset | Accuracy (%) | Recall (%) | F1-Score (%) |
|---|---|---|---|---|
| **GNN-ASE** | Eccentricity Defects | 92.5 | 88.3 | 86.7 |
| | Bearing Faults | 91.2 | 86.5 | 84.8 |
| | Bar Breakage Defects | 93.1 | 89.6 | 88.1 |
| **GNN-ASE@1** | Eccentricity Defects | 89.8 | 83.7 | 81.5 |
| | Bearing Faults | 87.6 | 82.1 | 80.4 |
| | Bar Breakage Defects | 90.2 | 85.4 | 83.9 |
| **GNN-ASE@2** | Eccentricity Defects | 92.5 | 88.3 | 86.7 |
| | Bearing Faults | 92.5 | 88.3 | 86.7 |
| | Bar Breakage Defects | 91.2 | 86.5 | 84.8 |
| **GNN-ASE@3** | Eccentricity Defects | 93.1 | 89.6 | 88.1 |
| | Bearing Faults | 86.3 | 80.8 | 79.0 |
| | Bar Breakage Defects | 89.1 | 84.1 | 82.3 |

## 4.6 Comparison with other works

In terms of anomaly detection, our approach with GNN-ASE surpasses the results of both Yu et al. [14] and Chisedzi et al. [42], particularly in the areas of electrical signal processing for defect detection, in comparison to other works. Yu et al. suggest a dynamic model-embedded framework for intelligent machine fault diagnosis in the absence of fault data. This framework is effective in detecting faults in new machines through the use of a digital twin approach. Nevertheless, the method's adaptability to a variety of datasets may be restricted by its reliance on dynamic modeling and parameter identification. Furthermore, their findings, although advantageous, are restricted to simulations of machine data that are fault-free. In contrast, GNN-ASE utilizes graph-based learning and integrates edge reweighting and severity estimation, which enables it to excel in the identification of anomalies with real-world data, resulting in higher accuracy, recall, and F1-scores.

In the same vein, Chisedzi et al.'s research is dedicated to the identification of broken rotor bar (BRB) defects in squirrel cage induction motors through the use of machine learning techniques such as deep learning and decision trees. Although they achieve competitive results with their Decision Tree Classification (DTC) method, particularly in terms of precision and accuracy for both loaded and unloaded motors, our GNN-ASE exhibits superior overall performance, particularly when applied to a broader range of defects beyond those limited to BRB. The dynamic nature of GNN-ASE, in conjunction with its capacity to process both time and frequency domain features, offers a more comprehensive solution for the detection of a diverse array of defects and the estimation of their severity, resulting in superior performance metrics across multiple datasets.

## 4.7 Interpretation of results and limitations

The effectiveness of graph neural networks in the context of anomaly detection in electrical systems is highlighted by the results obtained from the GNN-ASE model across a variety of datasets demonstrated in Figure 6, including eccentricity defects, bearing faults, and bar fracture defects. The model's capacity to detect anomalies and evaluate their severity is demonstrated by its high accuracy, recall, and F1-scores. The model's adaptability has been significantly improved by the implementation of dynamic edge reweighting, which enables it to refine the relationships between nodes and enhance the accuracy of detection in real-world scenarios. The model's robustness is enhanced by the combination of frequency-domain features and time-domain characteristics, which enable it to develop a more profound comprehension of signal patterns.

Nevertheless, it is imperative to recognize a number of constraints. Initially, the model's efficacy is contingent upon the quality of the input graph, which can be exceedingly susceptible to noise and incomplete data. The GNN-ASE's efficacy may be compromised if the graph structure fails to accurately reflect the relationships between nodes. The computational complexity of training GNNs, particularly when applied to large-scale datasets, is another constraint.

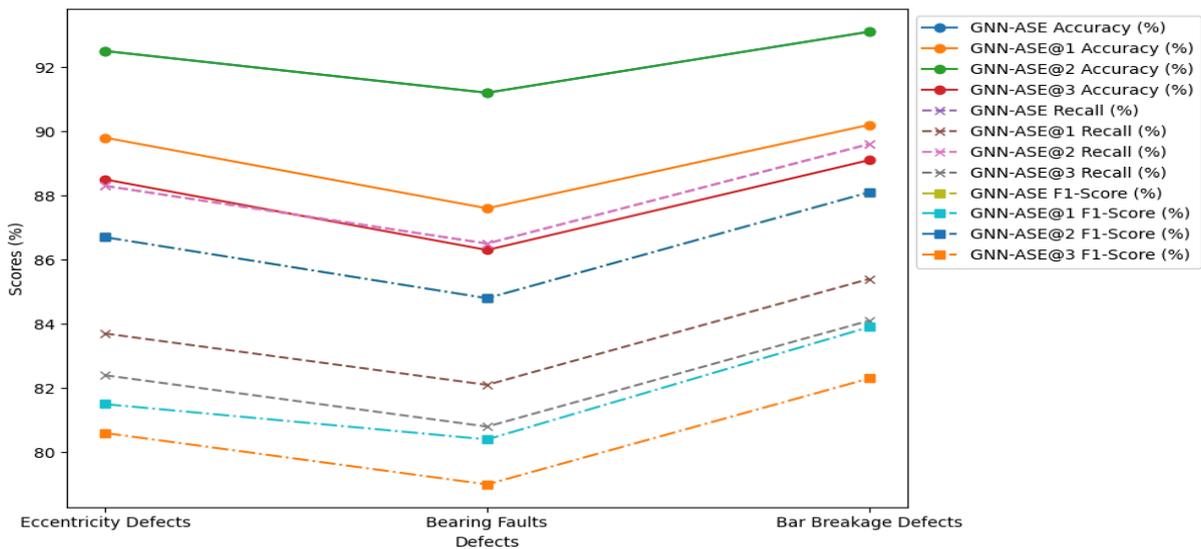

**Fig. 6 Performance Comparison of GNN-ASE Models**

## 5. Conclusion

The GNN-ASE model is a graph neural network-based model designed for anomaly detection in electrical systems, specifically focusing on faults like eccentricity defects, bearing faults, and bar breakage defects. The model captures temporal and frequency domain features and uses dynamic edge reweighting and anomaly severity estimation mechanisms for more nuanced understanding. The model demonstrated strong performance in accuracy, recall, and F1-score across multiple datasets, indicating its potential for industrial applications. It can handle imbalanced datasets and detect anomalies with varying severity levels. However, limitations in computational complexity and sensitivity to graph quality need to be addressed for scalability and generalization to diverse datasets. Future research will also focus on optimizing severity estimation and expanding the model's capabilities to detect a wider range of fault types.

**CRediT authorship contribution statement**
**Moutaz Bellah Bentrad:** Data curation, Methodology, Resources, Visualization, Writing - original draft. **Adel Ghoggal:** Data curation Supervision, Investigation - review & editing. **Tahar Bahi:** Supervision, Investigation, Writing - review & editing. **Abderaouf Bahi:** Methodology, Software, Validation, Writing - original draft.

**Conflict of interest**
The authors declare that they have no known competing financial interests or personal relationships that could have appeared to influence the work reported in this paper.

**Data availability**
Data will be made available on request.

**Acknowledgments**
The authors express their gratitude to the Algerian Ministry of Higher Education and Scientific Research (MESRS) for financially supporting this research.